\documentclass[conference,letter]{IEEEtran}
\IEEEoverridecommandlockouts
\usepackage{cite}
\usepackage{amsmath,amssymb,amsfonts}
\usepackage{algorithmic}
\usepackage{graphicx}
\usepackage{mathtools}

\usepackage{float}
\usepackage{set space}
\usepackage{enumitem}
\usepackage{wrapfig}
\usepackage[table]{xcolor}
\usepackage{adjustbox}

\usepackage{caption}
\usepackage[linesnumbered,ruled,vlined]{algorithm2e}

\SetCommentSty{mycommfont}
\usepackage{xcolor}
\def\BibTeX{{\rm B\kern-.05em{\sc i\kern-.025em b}\kern-.08em
    T\kern-.1667em\lower.7ex\hbox{E}\kern-.125emX}}

\usepackage{verbatim}

\usepackage{enumitem}
\renewcommand{\theenumi}{\Alph{enumi}}

\usepackage[caption=false, font=small]{subfig}

\newcommand{\red}[1]{\textcolor{red}{[#1]}}

\newcommand{\teal}[1]{\textcolor{teal}{#1}} 
\newcommand{\fref}[1]{Fig.~\ref{#1}}
\newcommand{\tref}[1]{Table~\ref{#1}}
\newcommand{\sref}[1]{Section~\ref{#1}}

\SetKwInput{KwInput}{Input}
\SetKwInput{KwOutput}{Output}
\SetKwInput{KwInitialize}{Initialize}

\begin{document}

\title{LightESD: Fully-Automated and Lightweight Anomaly Detection Framework for Edge Computing}

\author{\IEEEauthorblockN{Ronit Das}
\IEEEauthorblockA{\textit{Department of Computer Engineering} \\
\textit{Missouri University of Science and Technology}\\
Rolla, USA \\
rdkz8@umsystem.edu}
\and
\IEEEauthorblockN{Tie Luo$^*$}\thanks{$^*$Corresponding author.}\thanks{To appear in Proceedings of IEEE EDGE 2023, Chicago, July 2023.}
\IEEEauthorblockA{\textit{Department of Computer Science} \\
\textit{Missouri University of Science and Technology}\\
Rolla, USA \\
tluo@mst.edu}
}

\maketitle

\begin{abstract}
Anomaly detection is widely used in a broad range of domains from cybersecurity to manufacturing, finance, and so on. 
Deep learning based anomaly detection has recently drawn much attention because of its superior capability of recognizing complex data patterns and identifying outliers accurately. However, deep learning models are typically iteratively optimized in a central server with input data gathered from edge devices, and such data transfer between edge devices and the central server impose substantial overhead on the network and incur additional latency and energy consumption. To overcome this problem, 
we propose a fully-automated, lightweight, statistical learning based anomaly detection framework called LightESD. It is an on-device learning method without the need for data transfer between edge and server, and is extremely lightweight that most low-end edge devices can easily afford with negligible delay, CPU/memory utilization, and power consumption. Yet, it achieves highly competitive detection accuracy. Another salient feature is that it can auto-adapt to probably any dataset without manually setting or configuring model parameters or hyperparameters, which is a drawback of most existing methods. We focus on time series data due to its pervasiveness in edge applications such as IoT. Our evaluation demonstrates that LightESD outperforms other SOTA methods on detection accuracy, efficiency, and resource consumption. Additionally, its fully automated feature gives it another competitive advantage in terms of practical usability and generalizability.
\end{abstract}


\section{Introduction}
The inception of research in outlier detection can be dated to as early as 1852 when Benjamin Peirce came up with \textit{Peirce's Criterion} \cite{peirce1852criterion} to detect and remove outliers from numerical data. 
Since then, research in this area has grown remarkably and has now become ubiquitous in almost every domain, such as cybersecurity, transportation, manufacturing, finance, and computer networks.

With the rapid development of edge computing, a large number of applications that require real-time response have been moving to edge devices. The concern pertaining to cyber attacks, fault diagnosis, and other similar data analytics has also urged development of anomaly detection for edge computing. Traditional machine learning-based methods \cite{breunig2000lof, liu2008isolation} have seen a recent trend of being replaced by deep learning-based methods \cite{wang2022lightlog, geiger2020tadgan, sivapalan2022annet, an2015variational}, due to the latter's state-of-the-art (SOTA) performance in detection accuracy. However, deep learning models typically require intensive training and a large amount of data; as a result, a central server or cloud is often deployed which collects data from edge devices and then performs model training\cite{sivapalan2022annet}. This entails data transfer over the network and adds substantial network traffic and overhead, as well as incurs large delay. An alternative is to train the model offline at a central server using all the historical data, and then deploy the model at edge devices for inference only. However, this approach is not able to keep up with new data and can lead to the problem of {\em concept drift}\cite{lu2018learning}.

In this paper, we propose a fully-automated, lightweight, statistical learning-based anomaly detection framework called LightESD, for detecting anomalies directly at the edge site. It is extremely lightweight with little resource consumption and little training overhead that almost all edge devices can afford (we have quantified these in our evaluation). A salient feature of LightESD is that it is a {\em weight-free}, unsupervised model, meaning that it stores no weights. This enables it to {\em auto-adapt to any data} to learn the underlying patterns on the fly in a fully unsupervised manner, without the need for manual {\em pre-processing or hyperparameter-tuning} to ``match'' any specific dataset. Thus, it is much favorable for practical adoption, and has good generalizability over different data.

Another important feature of LightESD is that it is {\em non-parametric}, meaning that it does not make any distributional or functional-form assumptions of the observed (original) data, whereas many other statistic approaches do. This also contributes to its good generalizability. Note that {\em being non-parametric is not equivalent to being weight-free}; for example, SVM \cite{cristianini2000introduction, cortes1995support} with RBF-kernel is non-parametric but has weights.\footnote{Support vector coefficients (which are dual coefficients) are learnt from training data and stored in the memory for making predictions. Hence whenever we have another dataset, the model needs to be retrained and the coefficients (weights) need to be replaced. However, our proposed approach auto-adapts to any underlying data to make predictions on the fly, without explicit training or the need for storing any coefficients/weights.} 



This paper focuses on time series which pervades in numerous application domains (such as Edge) that have temporal properties. The main contributions of this paper are:
\begin{enumerate}[label=\arabic*)]

    \item We propose LightESD, an anomaly detection framework for time series, that realizes on-device learning and suits deployment on edge devices. LightESD is weight-free, non-parametric, unsupervised, and can auto-adapt to perhaps any univariate time series regardless of the underlying distribution, seasonality, or trend, without manual pre-processing or hyperparameter-tuning.
    
    \item We propose a new evaluation metric, {\em ADCompScore}, that allows for comparison of edge anomaly detection models in a holistic manner. To the best of our knowledge, this is the first attempt to develop a new metric to understand not only the anomaly detection performance but also the computational power and resource usage, using a single numeric value which allows for quick decision-making regarding the feasibility of deploying anomaly detection algorithms on edge devices.
    \item We evaluate LightESD using both synthetic and real-world datasets and demonstrate its superior overall performance compared to SOTA methods, in terms of both anomaly detection and feasibility for on-device training and edge deployment.
\end{enumerate}

The rest of this paper is organized as follows. Section II discuss related work as the background. Section III presents the proposed LightESD approach. Section IV describes our experiments and comprehensive evaluation of the proposed method in comparison with other methods. Section V concludes with future directions.

\section{Related Work}\label{sec:relwk}
{\bf Statistical methods for anomaly detection.}
These have been the go-to approaches for a long time, such as ARIMA  and linear regression. While they require minimal effort for the model to learn from data, they hinge on the \textit{normality} assumption that the underlying data must conform to a \textit{Gaussian} or \textit{Gaussian-like} distribution \cite{chandola2009anomaly}, which often does not hold in real-world data or cannot characterize multi-modal distribution encountered in some datasets\cite{hochenbaum2017automatic}. This makes such models, in their standalone form, unsuitable for detecting anomalies in real-world data.
{\bf Machine Learning and Deep Learning based anomaly detection.}
Some of the commonly used machine learning-based methods include distance based techniques \cite{zhu2020knn}, density-based techniques \cite{breunig2000lof}, tree-based techniques \cite{liu2008isolation}, Bayesian networks \cite{mascaro2014anomaly}, and clustering techniques \cite{pu2020hybrid}. Most machine learning-based anomaly detection methods have been superseded by deep learning-based approaches owing to the latter's much better anomaly detection performance. Indeed, recent advances in deep learning (DL) have created a hype of using DL in nearly all tasks including anomaly detection. Unsupervised DL architectures include those based on deviation networks \cite{pang2019deep}, adversarial learning such as f-AnoGAN~\cite{schlegl2019f} and TadGAN~\cite{geiger2020tadgan}, an other variants. While deep learning-based methods can achieve good performance, they often come with a large memory and computational footprint, which can pose a bottleneck for deployment on edge devices.

{\bf Anomaly Detection for the Edge.}
While there exist several papers that discuss the hardware-implementation of neural network-based anomaly detection approaches like ANNet\cite{sivapalan2022annet}, LightLog\cite{wang2022lightlog}, and others, all of them adopt the approach of training a model at a central server and then deploying (``implementing'') it on the device hardware. On the other hand, ONLAD \cite{tsukada2020neural} proposes to develop an OS-ELM\cite{liang2006fast} based approach to detect anomalies, which can be directly trained and deployed at an edge site. Since \cite{tsukada2020neural} attempts to solve the same challenge as we do, we compare it with our proposed approach in \sref{subsec:result}.

Some of the recent works, like that of Bayesian Random Vector Functional Link AutoEncoder with Expectation Propagation (EPBRVFL-AE)\cite{odiathevar2022bayesian}, take a distributed approach to training their proposed anomaly detection schemes. While taking a distributed or federated learning-based approach might seem intuitive, such a direction can increase the overall communication overhead in a network, thereby degrading the network's efficiency. Moreover, such approaches still require a central server and a communication network which are not always available especially in rural or challenging environments, which is where our work fits in.

\section{The LightESD Framework}\label{sec:anom_detect}
LightESD stands for {\em Lightweight Extreme Studentized Deviate} test.
In a nutshell, it works as follows. A time series $Y_{t}$ can be decomposed additively as
\begin{equation}\label{eq:add_decompose}
    Y_{t} = T_{t} + \sum_{i = 1}^{k}S^{i}_{t} + R_{t}
\end{equation}
where $T_{t}$, $S^{i}_{t}$ and $R_{t}$ represent the {\em trend}, the $i$-th {\em seasonal}, and the {\em residual} components, respectively, and there are a total of $k\ge 1$ seasonal components. LightESD first detects all the different seasonal periodicities in the input time series, i.e., to determine $S^i_t$, where it uses {\it Welch's Periodogram Method} with a PSD-locating technique improved by us (\sref{sec:period}). 
Second, it extracts the trend component $T_t$ using {\it RobustTrend} and {\it FastRobust-STL} decomposition methods, and then removes both the detected trend and seasonality components, to extract the residual $R_t$ (\sref{sec:decomp}). 
Third, LightESD detects anomalies based on the residual---which is a statistically correct (because the residuals follow a normal or approximately-normal distribution) and much more reliable method than detecting based on original signals---using a generalized ESD test with our improvement in robustness
(\sref{sec:detect}). Note that, our choice of the above methods  is based on careful contemplation and trial experiments. We explain our choice below when we describe those methods.

\subsection{Improved Periodicity Detection}\label{sec:period}
Time series data often exhibit a recurring pattern at regular time intervals like weekly, monthly, or yearly, which is called the \textit{seasonality} of a time series. We detect seasonality by first computing a {\em periodogram}, 
which is an estimate of the power spectral density (PSD) of a signal, and then analyzing the periodogram to find out the dominant frequencies that generate the highest PSD estimates above a designated threshold. There are both parametric and non-parametric methods to estimate PSD, 
where parametric methods assume an underlying data distribution which may not hold. Therefore, we take a non-parametric approach.

\subsubsection{Welch's Periodogram Method}

Among the few non-parametric methods that can compute the Periodogram for a time series,
we choose Welch's method\cite{welch1967use} as it outputs the most reliable PSD estimates, even from noisy data. This method first splits the signal into $K$ data segments of length $L$, which are overlapped by $D$ points ($D/L$ ranges from 0 to $0.5$). This can reduce the effect of noise on PSD estimation, unlike {\it Bartlett's method} which has no overlapping. 
Second, each of the $K$ overlapping segments is applied a time window that is either quadratic ($1-t^2,\, t\in[-1,1]$) or triangular ($1-|t|,\, t\in[-1,1]$). Third, a periodogram is generated for each window by first computing the discrete Fourier transform (DFT)~\cite{kurtz1985algorithm} and then the squared magnitude of the DFT output. Finally, the individual periodograms are averaged to reduce the variance of individual power measurements.

\begin{algorithm}[t]
\caption{Improved Period Detection}
\label{algo:period}
\DontPrintSemicolon
  \KwInput{$Y$: Time series}
  \KwOutput{$prd$: Array of detected periods}
  

  \For{$i = 1$ to $100$}
  {
    $Y' \gets {\tt\small permutation}(Y)$\\
    $freq, pow \gets {\tt\small Welch}(Y')$\\
    $p_{max} \gets {\tt\small max}(pow)$\\
    $max\_power.{\tt\small add}(p_{max})$\\
  }
  $max\_power \gets {\tt\small sort}(max\_power, {\tt\small ascending=true})$ \\
  $index \gets 0.99 \times {\tt\small len}(max\_power)$\\
  $thresh \gets max\_power[index]$\\
  $freq, pow \gets {\tt\small Welch}(Y)$ \\

  $prd \gets -1$, $temp\_psd \gets -1$\\

  \For{$j = 1$ to ${\tt\small len}(pow)-1$}
  {
    \If{ ($pow[j]>{thresh}$) and ($pow[j] > pow[j-1]$) and ($pow[j] > pow[j+1]$) }
    {
      \If{ ($pow[j] > {temp\_psd}$)} 
      {
        $prd.{\tt\small add} (\left\lfloor\dfrac{1}{ freq[j] }\right\rfloor)$ \\
        $temp\_psd \gets pow[j]$\\
        
      }
      
    }
  }

  \If{($prd == -1$)}
  {
    $prd \gets 1$ \tcp{Nonseasonal}
  }
  
  \KwRet $prd$
\end{algorithm}
\subsubsection{Improved Periodicity Detection}

From the obtained periodogram, we need to find the significant PSD peaks which correspond to the different periods (seasonalities), if they exist. Although Welch's method reduces noise to some extent via overlapping segments, it is fairly primitive and the remaining noise in time series still creates many spurious PSD peaks in the resulting periodogram. Therefore, we improve a periodicity detection approach {\it AutoPeriod} \cite{vlachos2005periodicity} 
to filter out those spurious peaks. Specifically, we first permute the original time series for 100 times and compute the periodogram for each permuted sequence using the Welch's method. In each iteration, the random permutation destroys all the temporal correlations as well as the second and higher-order central moments of the original time series, and thereby convert the time series into pure noise. Among all the PSD values generated from each noise series using the {\em Welch's} method, we take the maximum PSD value and add it into a vector. After that, the vector containing the $100$ PSD values is ordered in an ascending manner, and the PSD value at the $99$th index is selected as the threshold value (basically it calculates the $99$-th percentile). The reason for choosing this particular value as the threshold is that $99\%$ of the PSD values come from the permuted version of the original time series (noise), which always lie below the $99$-th percentile value (also known as the threshold PSD value). So any PSD estimate lying above this threshold is significant, at $99\%$ confidence, while those lying below are just PSD estimate from pure noise which we disregard. 

Our contribution to this period detection algorithm is that, unlike \cite{vlachos2005periodicity} which takes {\em all} the PSD values and has a high space complexity of $\mathcal{O}(N)$, our proposed periodicity detection method selects significant peaks only (since a peak represents a cyclic event) which has a space complexity of merely $\mathcal{O}(1)$. In other words, as the length of the time series increases, the space consumed by \cite{vlachos2005periodicity} increases approximately proportionally as it considers all the significant values whereas our method remains constant. For time complexity, both \cite{vlachos2005periodicity} and our approach are at the same level $\mathcal{O}(N\log N)$, due to the usage of Fast Fourier Transform.

\begin{figure}[ht]
    \centering
  \subfloat[Seasonal periods are correctly detected.]
  {\includegraphics[width=.9\linewidth]{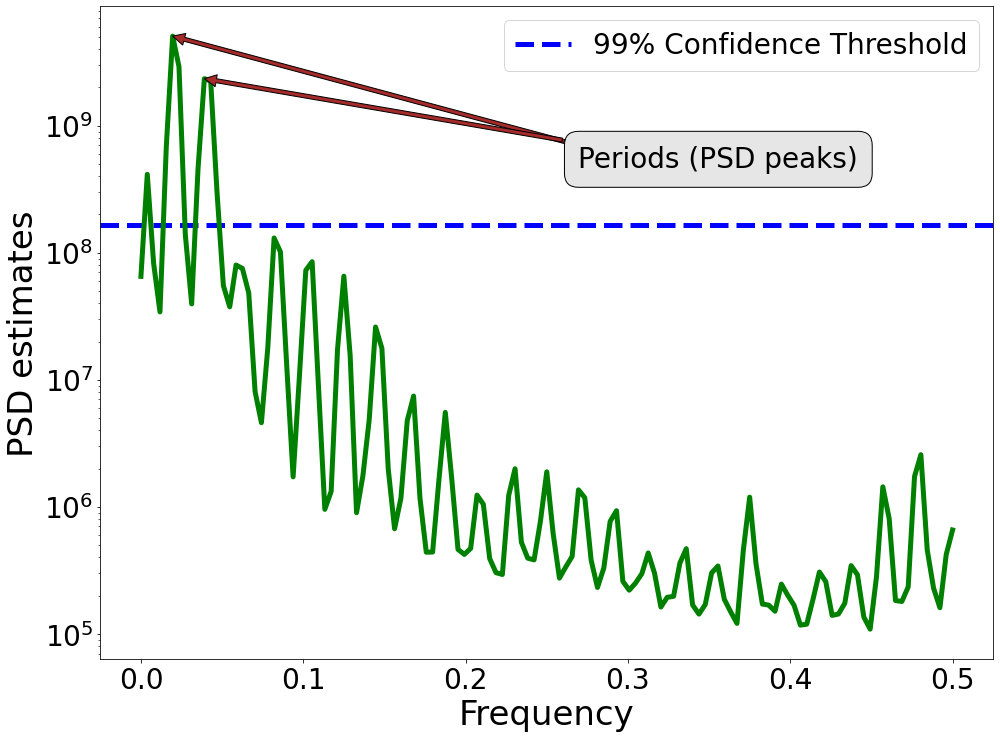} \label{fig:period}} 
\vfill
  \subfloat[Non-seasonality also detected (not fooled by noise).]
  {\includegraphics[width=0.9\linewidth]{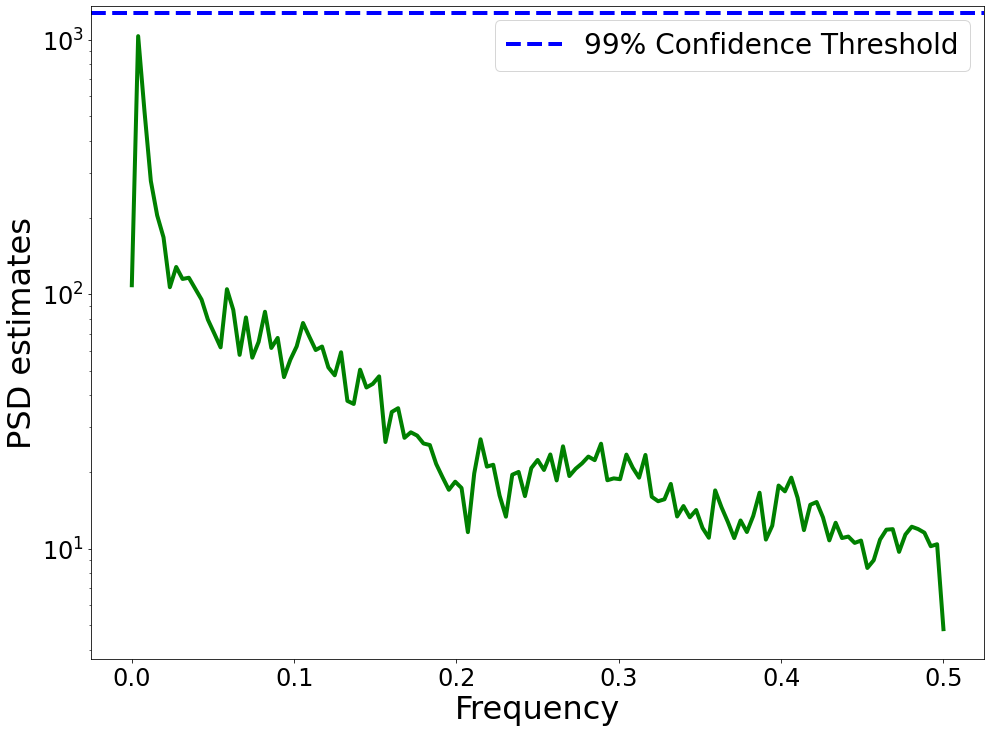}}\label{fig:aperiod}
  \caption{Illustration of our periodicity detection result on NAB \cite{lavin2015evaluating}: (a) NYC Taxi dataset (seasonal), (b) UPS Tweets dataset (nonseasonal).}

  \label{fig:period_det} 
\end{figure}

The above procedure is presented in Algorithm~\ref{algo:period} and the outcome is illustrated in \fref{fig:period_det} using the Numenta Anomaly Benchmark (NAB) dataset \cite{lavin2015evaluating}. We have considered other approaches that detect periodicities, such as {\it AutoAI-TS}\cite{shah2021autoai}, where the latter uses {\em pre-defined} schemes only (like hourly / daily / weekly periodicities) to detect periodicities (seasonalities) in the time domain (whereas we detect in the frequency domain). This has the limitation of not being able to detect periods that are {\em not} in the manually pre-defined set. 

\subsection{Residuals Extraction}\label{sec:decomp} 
To extract the residuals $R_t$, we use {\it RobustTrend}  \cite{wen2019robusttrend} if the time series is nonseasonal (aperiodic) where our Algorithm~\ref{algo:period} will return $prd=1$. Otherwise, if the series is seasonal, we use \textit{FastRobust-STL} \cite{wen2020fast}
to extract both the seasonality and the residuals. There are four reasons why we choose {\it RobustTrend}  \cite{wen2019robusttrend} for decomposing aperiodic time series and why \textit{FastRobust-STL} \cite{wen2020fast} for seasonal series: (1) these two methods can model both abrupt and slow-changing trend components, (2) they are robust to outliers and noise, (3) {\it FastRobust-STL} is able to handle multiple seasonalities and can decompose much faster than other comparable methods, and (4) both approaches are free from assumption of any particular distribution and are non-parametric.  

\subsubsection{Nonseasonal Series}
In this case, we use {\it RobustTrend} to de-trend and extract residuals. In order to extract the trend from the time series in a reliable way, it is important to mitigate the negative impact of noise and outliers. This is achieved by minimizing the \textit{Huber Loss} of the residual signal, combined with first and second order difference regularization, as follows:
\begin{align}\label{eq:fit}
    \arg\min_{\mathbf{t}  } h_{\gamma}(\mathbf{y} - \mathbf{t}) + \lambda_{1}{||\mathbf{D}_{(1)}\mathbf{t}||}_{1} + \lambda_{2}{||\mathbf{D}_{(2)}\mathbf{t}||}_{1}\hspace{1pt}
\end{align}
where $h_{\gamma}(.)$ is the \textit{Huber Loss}, $\mathbf{D}_{(1)}$ is the first-order difference matrix, $\mathbf{D}_{(2)}$ is the second-order difference matrix, and $\lambda_{1}$ and $\lambda_{2}$ control the amount of the regularization.

The optimization problem \eqref{eq:fit} can be solved using \textit{Alternate Direction Method of Multipliers} (ADMM) based on \textit{Majorization-Maximization} \cite{sun2016majorization}, to estimate the trend $\mathbf{t}^* = T_t$. Then the residuals are extracted by $R_t = Y_t - T_t$.

\subsubsection{Seasonal Series}
In this case, we use {\it FastRobust-STL} \cite{wen2020fast} which first extracts the trend using a robust sparse model. In order for accurate estimation of the trend component, first we need to remove noise and the adverse effect of the different seasonalities (note that this is different from estimating the different seasonal components), as well as to take into account possible outliers. To remove noise, a \textit{bilateral filter} \cite{thompson2014empirical} is used to denoise the time series.
To remove the influence of seasonal components on trend extraction, a seasonal differencing operation is carried out, which refers to the difference between the outputs of the bilateral filter for every timestep $t$ and that for $t - T$, where $T$ is the largest seasonal period (i.e., $T=\max(prd)$) in the time series. 

Then, the trend is extracted by formulating an objective function that minimizes the least absolute deviation (LAD) of the smoothed (de-noised) and seasonal-differenced signal, in order to make the trend extraction robust to large outliers; the objective is further combined with $L_1$-regularizations to capture both abrupt and slow trend changes.

Then, after extracting and removing the trend, in order to estimate the multiple seasonal components, an improved version \cite{wen2020fast} of a non-local seasonal filter \cite{wen2019robuststl} is used to extract the different seasonal components present in the time series.

Finally, the residual component is extracted by subtracting the trend and the multiple seasonal components from the original time series, as $R_{t} = Y_{t} - T_{t} - \sum_{i = 1}^{k}S^{i}_{t}$.

\subsection{Anomaly Detection based on Residuals}\label{sec:detect}

The rationale that we perform anomaly detection based on residuals instead of original data, is that such residuals as we extracted above, are unimodal and follow an approximately Gaussian distribution. This stems from the fact that trend-adjusted series (time series with its trend removed) are {\em stationary} and have been empirically shown to follow an approximately-Gaussian distribution \cite{hochenbaum2017automatic}. This is an important property that makes our anomaly detection much more accurate than other anomaly detection methods when the original data distribution is not Gaussian-like or unimodal. For example, using any real-world data (which may have multiple mode(s)) directly for detecting anomalies, can result in many of the anomalies not being detected at all\cite{hochenbaum2017automatic}.

A classical approach to outlier detection is to conduct a {\em hypothesis test} at a certain significance level to decide whether to reject the null hypothesis (a data point in question is {\em not} an outlier). Some of the most common statistical methods including Pierce's Criterion\cite{ross2003peirce} 
and the Tietjen Moore Test\cite{tietjen1972some} take this approach. However, these methods have various drawbacks such as being not scalable to large datasets or requiring prior knowledge of the exact number of outliers in the data. In LightESD, we propose a method based on an improvement to the generalized Extreme Studentized Deviate (ESD) test \cite{rosner1983percentage}.

\subsubsection{Generalized Extreme Studentized Deviate Test}\label{lab:grubb}

The ESD test \cite{rosner1983percentage} enhances the standard Grubb's Test \cite{grubbs1950sample} in that ESD can find up to a user-specified maximum, say $a_{max}$, of outliers in a series of data points, whereas Grubb's Test can only find a single outlier. 
However, both ESD and Grubb's Test assume normality of the input data which often does not hold, while our method does not make that assumption. ESD runs $a_{max}$ iterations and, in each iteration, it tests if a single outlier exists by comparing a test statistic $R$ with a critical value $\lambda$. The test statistic $R$ is defined as
\begin{equation}\label{eq:grubb}
    R = \max_{i} \left(\frac{|Y_{i} - \mu|}{\sigma}\right),\; i=1,...,n
\end{equation}
where $\mu$ and $\sigma$ are the mean and standard deviation, respectively, of the input data $Y$ of length $n$. The critical value $\lambda$ is defined with respect to a significance level $\alpha$, as
\begin{equation}\label{eq:crit}
    \lambda = \frac{t_{n-l-2,p} \times (n-l-1)}{\sqrt{(n-l) \times (t_{n-l-2,p}^{2} + n -l - 2)}}
\end{equation}
where $p = 1 - \frac{\alpha}{2} \times(n - l)$, $l$ is the iteration index ranging from 0 to $a_{max}-1$, and $n$ is the current length of the series. The value of $t_{(.,.)}$ can be looked up in the {\em two-tailed T-Distribution table}.
If $R>|\lambda|$, we reject the null hypothesis that $Y_{i^*}$ is not an outlier, where $i^*$ is the index $i$ that yields $R$ as in \eqref{eq:grubb}. Then we remove this outlier $Y_{i^*}$ from $Y$, decrement $n$ by 1, and move to the next iteration (increment $l$ by 1). Otherwise, no outlier is detected in this iteration and the algorithm continues to the next iteration until $l$ reaches $a_{max}$. 

\subsubsection{Improvement to ESD}\label{subsec:robust}
Besides that we do not make the normality assumption of observed data that ESD makes when it operates on the observed data in a standalone form, we also identify that ESD test is vulnerable to deviant points: 
the test statistic $R$ used by ESD in (\ref{eq:grubb}) will break down when the series $Y$ contains a small fraction of, or even just a single, very large outlier. 
To address this issue, we redefine the test statistic $R$ as
\begin{align}\label{eq:robustR}
    R_{robust} &= \max_{i} \left(\frac{|Y_{i} - \text{median}(Y)|}{S(Y)}\right), \; i=1,..,n\\
S(Y) &= \text{median}_{i}\left(\text{median}\left|Y_{i} - Y\right|\right).
\end{align}
That is, we replace $\mu$ and $\sigma$ as in \eqref{eq:grubb} with median and a {\em robust estimator $S$}\cite{rousseeuw1993alternatives}, respectively. 
This is important and implies that the {\em finite sample breakdown point} \cite{iglewicz1993volume}
will now change from $1/(n+1)$ to $\lfloor \frac{n}{2} \rfloor / n$. What this means is that our method is able to tolerate $50\%$, in the asymptotic case, of all the values to be arbitrarily large. Therefore, our method is far more robust than the original ESD since outliers can never exceeds $50\%$ (otherwise they would not be called ``outliers'').

Moreover, it is also worth noting that we choose the robust statistic $S$ over {\em Median Absolute Deviation} (MAD) as used by \cite{hochenbaum2017automatic}, even though both can achieve asymptotic $50\%$ toleration of outliers. The rationale is that the statistic $S$ does not make the assumption of {\em symmetric distribution} that MAD requires \cite{huber2011robust}, and $S$ is more efficient than MAD when dealing with Gaussian distributions \cite{rousseeuw1993alternatives}.

\subsection{Putting All Together}
The entire anomaly detection process is summarized in Algorithm \ref{algo:3}. 
At Line \ref{l:amax}, $a_{max}$ is set to $10\%$ of the total number of data points since a real-world dataset typically contains less then 4-5\% of anomalies. Importantly, note that this value is highly {\em insensitive} to different datasets since it is an {\em upper bound} to the number of anomalies rather than the exact quantity. The runtime complexity of Algorithm \ref{algo:3} is $\mathcal{O}(N\log N)$, owing to Algorithm \ref{algo:period}. The space complexity of Algorithm \ref{algo:3} is $\mathcal{O}(N)$, since in the worst case scenario, the detected anomalous time indices can grow as a fraction of the length of the entire time series being analyzed.

\begin{algorithm}[h]
    \DontPrintSemicolon
  \KwInput{$Y$: Time Series} 
  \KwOutput{$anomalies$: the corresponding indices}
  
  \KwInitialize{$outlier\_index \gets$ []}

  ${period} \gets {\tt\small ImprovedPeriodDetection}(Y)$ \tcp{Algorithm 1}
  
  \If{${period} == 1$}
  {
    ${residual} \gets Y - {\tt\small RobustTrend}(Y)$\\
  }
  \Else 
  {
    ${residual} \gets {\tt\small Fast RobustSTL}(Y, period)$\\
  }
  
  $a_{max} \gets 0.1\times{\tt\small len}(Y)$ \label{l:amax}\\
  ${outliers} \gets {\tt\small ImprovedESD}(\alpha = 0.05, residual, a_{max})$\\ 
  
  ${outlier\_index} \gets{outliers.index}$\\ 
  
  \If{(${outliers}[1] == {\tt\small True}$) {and} (${outliers}[2] == {\tt\small False}$) \label{l:beg}}
  {
    ${outliers}[1] = {\tt\small False}$
    
    ${outlier\_index}.{\tt\small pop}(1)$
  }
  
  \If{(${{outliers}[n] == {\tt\small True}}$) {and} (${{outliers}[n-1]} == {\tt\small False}$)}
  {
    ${outliers}[n] = {\tt\small False}$
    
    ${outlier\_index}.{\tt\small pop}(n)$
    \label{l:end}
  }
  
  

  ${anomalies} \gets {outlier\_index}$

  \KwRet $anomalies$

\caption{LightESD: The complete procedure}
\label{algo:3}
\end{algorithm}

\section{Performance Evaluation}\label{sec:eval}
\subsection{Experiment Setup}
\begin{figure}[t]
    \centering
  {\includegraphics[width=0.9\linewidth]{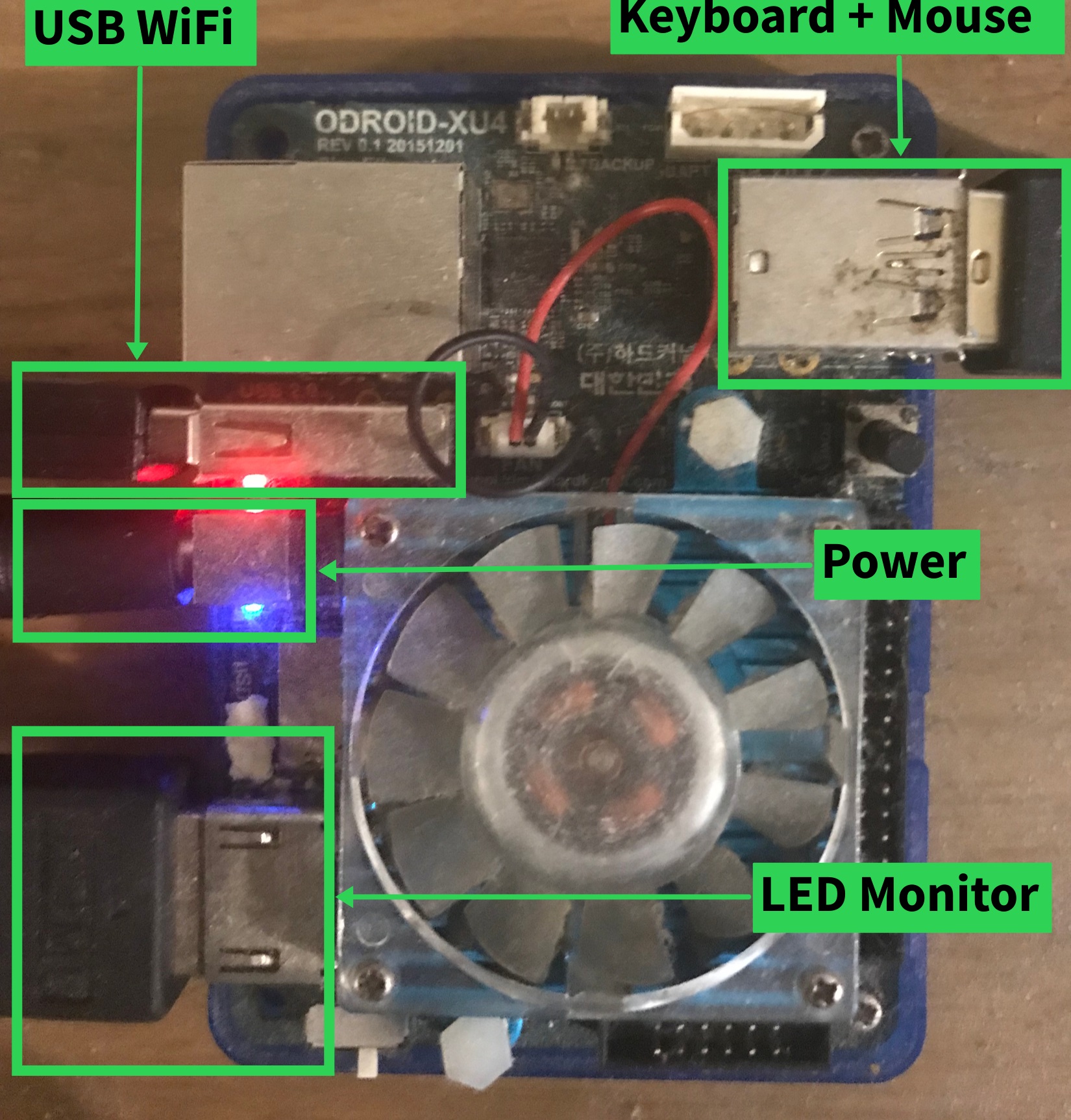} }
    \caption{The Odroid experimental setup including peripherals.}
    \label{fig:odroid}
\end{figure}

We used a single board computer, Hardkernel Odroid XU4 (OS: Ubuntu $22.04$, CPU: Samsung Exynos $5422$ Cortex A$15$ $2.0$ GHz and Cortex-A$7$ Octacore CPU, RAM: $2$ GB LPDDR$3$ PoP stacked), as our experimental machine to emulate an edge device, that is capable of on-device training and deployment of anomaly detection models. See \fref{fig:odroid} for the setup.

{\bf Datasets.} We evaluate our proposed LightESD framework on both synthetic and real-world datasets.
 We carefully synthesize two types of datasets that emulate different real-world univariate time series, as follows:
\begin{enumerate}[label=\arabic*),topsep=0pt]
    \item {\it Seasonal data (with both trend and seasonality)} ($\mathcal{STD}$):
\begin{align}\label{eq:seas}
    Y_t = \kappa \cdot t^{2} + \beta \cdot \sin(\frac{2\pi t}{30.5}) + \gamma \cdot \epsilon_{t}
\end{align} 
where $\kappa,\beta,\gamma$ are drawn randomly from [0.001,0.01], [1.3e+04, 1.5e+04], and [1.5e+03, 3.0e+03], respectively, and $\epsilon_{t} \sim N(0, 1)$ is gaussian white noise. The period is 30.5 (days) to emulate a monthly seasonality.


    \item {\it Random Walk (non-seasonal, no trend}) ($\mathcal{RW}$):
\begin{equation}
    Y_{t} = Y_{t-1} + \epsilon_{t}
\end{equation}
with $Y_{0} = 1.0$ and $\epsilon_{t} \sim \mathcal{N}(0, 1)$ is the gaussian white noise, where the current value of the series is only dependent on the previous timestep value. Note that random walk is different from Gaussian white noise.
\end{enumerate}

\begin{table}[ht]
    \centering 
    \caption{Injected anomalies by type (percentages are w.r.t. the total no. of anomalies)}
    \label{tab:anom}
  
\resizebox{0.9\linewidth}{!}{
  \begin{tabular}{|c|c|c|c|c|}
    
    \hline
    Dataset & \#Anom. & Spikes & Dips & Coll. Anom. \\
    \hline
    $\mathcal{STD}$ & $7$ & $43\%$ & $28.5\%$&$28.5\%$\\
    \hline
    $\mathcal{RW}$ & $9$ & $44\%$ & $44\%$&$12\%$\\
    \hline
\end{tabular}
}
\end{table}

Each synthetic dataset has 5,000 data points. The outliers are then injected with varying magnitudes and types  
as follows. The magnitudes randomly vary from $0.5\sigma$ to $6.0\sigma$ as in \cite{hochenbaum2017automatic}, where $\sigma$ is the standard deviation of the dataset concerned. The types and their respective quantities are given in Table \ref{tab:anom}. The positions of anomalies are random.

For {\bf real-world datasets}, we use the Numenta Anomaly Benchmark dataset \cite{lavin2015evaluating} (the \textit{realKnownCause} category) and the Yahoo Anomaly Detection dataset\cite{laptev2015s5} (the \textit{A1 Benchmark} category). These univariate time series datasets incorporate complex dynamics which is hard to emulate using synthetic data, thereby allowing us to evaluate our model's overall performance more comprehensively.

{\bf Baselines.} We compare our proposed model with SOTA neural network models dedicated to edge, as well as  with popular time-series anomaly detectors. We categorize the baseline models into (a) {\em Neural Network-based approaches}, and (b) {\em Machine Learning-based approaches}, and describe them as follows:
\begin{enumerate}[label*=\alph*.]
    \item 
    {\em Neural Network-based approaches.} We consider two SOTA approaches that employ neural network-based detection architecture.
    \begin{enumerate}[label*=\arabic*)]
        \item {\scshape EPBRVFL-AE : } {\em Bayesian Random Vector Functional Link AutoEncoder with Expectation Propagation} \cite{odiathevar2022bayesian} is the latest model for detection of anomalies at the edge in a distributed manner and is a single layered neural network. We re-implemented {\scshape EPBRVFL-AE} \cite{odiathevar2022bayesian} without the \textit{expectation propagation} as originally proposed in \cite{odiathevar2022bayesian}, because we are only interested in detecting anomalies on a single edge device without inter-node communication overhead. From here onwards, we refer to this re-implemented {\scshape EPBRVFL-AE} simply as {\scshape BRVFL-AE}. All the hyperparameters are set as proposed in the paper\cite{odiathevar2022bayesian}.

        \item {\scshape ONLAD \cite{tsukada2020neural}:} A single-layer feed-forward neural network model that was implemented directly on edge hardware for anomaly detection. We implement three different versions of {\scshape ONLAD}, each with $16, 64,$ and $128$ neurons in the hidden layer, respectively. The performance measures for ONLAD, as discussed in the later sections, are averaged over all the  three models.
    \end{enumerate}    

    \item {\em Machine Learning-based approaches.} We consider three commonly used machine learning-based approaches and describe them briefly below.

    \begin{enumerate}[label*=\arabic*)]
        \item {\scshape One-Class SVM (OC-SVM):}  First introduced by \cite{scholkopf1999support} as a one-class classifier based on {\em support vector machine}, mostly used for {\em novelty detection}. Its robust performance of detecting {\em novelties} (anomalies), along with very easy-to-use approach has made it one of the top choices for anomaly detection.

        \item {\scshape Local Outlier Factor (LOF):} It is a density-based machine learning approach, developed by \cite{breunig2000lof}. This approach is lightweight, along with good performance on detecting anomalies, thereby making it a suitable choice for many IoT/edge applications.

        \item {\scshape Isolation Forest:} This is a tree-based machine learning aproach, first proposed by \cite{liu2008isolation}. A high anomaly detection performance, with very few parameters to save, makes it an overall excellent candidate for anomaly detection problems.
    \end{enumerate}   
\end{enumerate}
The hyperparameters for the machine learning-based approaches (OC-SVM, LOF, and IF) are set to their default values as by the {\scshape scikit-learn} library. \par 
While other SOTA deep learning-based anomaly detection approaches do exist, like ANNet\cite{sivapalan2022annet} and LightLog\cite{wang2022lightlog}, they are not developed to be trained directly at an edge site (device), due to their computational and space complexities. Hence, these models are not considered as baselines in our evaluation.

The baseline methods as well as our proposed LightESD are all implemented in {\scshape Python 3.10}.  Moreover, we compare the baselines against our proposed model at $95\%$ (LightESD-1; $\alpha = 0.05$) and $99.9\%$ (LightESD-2; $\alpha=0.001$) confidence levels, where $\alpha$ is the significance level introduced in Section \ref{lab:grubb}.

{\bf Evaluation Metrics.} For a good understanding of model performance from different perspectives, we evaluate {detection performance} ({Precision}, Recall, and {$F_{1}$-score}), {generality}, {latency}, {resource utilization}, and {power consumption} of all the different models. 

\begin{table*}[!ht]
  \centering
  \caption{Detailed Performance Comparison on Detection}
  \label{tab:perf}
\resizebox{1.0\linewidth}{!}{
      \begin{tabular}{|c|c|c|c|c|c|c|c|c|c|c|c|c|c|c|c|c|c|}
        \hline
        &\multicolumn{3}{|c|}{$\mathcal{STD}$} & \multicolumn{3}{|c|}{$\mathcal{RW}$} & \multicolumn{3}{|c|}{NAB} & \multicolumn{3}{|c|}{Yahoo}&\multicolumn{2}{|c|}{\textbf{Generality}}\\
        \cline{2-15}
        \textbf{Models}&\textbf{Prec.}&\textbf{Rec.}&\textbf{F1}&\textbf{Prec.}&\textbf{Rec.}&\textbf{F1}&\textbf{Prec.}&\textbf{Rec.}&\textbf{F1}&\textbf{Prec.}&\textbf{Rec.}&\textbf{F1}&\textbf{Mean}&\textbf{CV}\\
        \hline
        OC-SVM & $0.51$ & $0.71$ &$0.59$&$0.5$ & $0.72$ & $0.59$&$0.50$ & $0.64$ & $0.56$&$0.52$ & $0.69$ & $0.59$&$0.58$&$0.03$\\
        \hline
        LOF & $0.62$ & $0.65$ & $0.63$&$0.58$ & $\mathbf{0.98}$ & $0.73$&$0.52$ & $0.7$ & $0.6$&$0.61$ & $0.77$ & $0.68$&$0.66$&$0.1$\\
        \hline
        Iso. For. & $0.50$ & $0.47$ & $0.48$&$0.50$ & $0.75$ & $0.6$&$0.50$ & $0.73$ & $0.59$&$0.54$ & $0.83$ & $0.65$&$0.58$&$0.12$\\
        \hline
        
        ONLAD & $0.70$ & $0.72$ & $0.71$&$0.66$ & $0.69$ & $0.67$&$0.8$ & $0.73$ & $0.76$&$0.75$ & $0.72$ & $0.73$&$0.72$&$0.06$\\
        \hline
        BRVFL-AE &$0.67$ & $0.76$ & $0.71$&$0.7$ & $0.88$ & $0.78$&$0.79$ & $0.8$ & $0.79$&$0.8$ & $0.73$ & $0.76$&$0.76$&$0.05$\\
        \hline
        \rowcolor{yellow!20}
        LightESD-1 & $0.77$ & $\mathbf{0.82}$ & $0.79$&$0.70$ & $0.97$ & $0.81$&$0.77$ & $\mathbf{0.83}$ & $0.80$&$0.80$ & $\mathbf{0.88}$ & $0.84$&$0.81$&$\mathbf{0.02}$\\
        \hline
        \rowcolor{green!20}
        LightESD-2 & $\mathbf{0.89}$ & $0.78$ & $\mathbf{0.83}$&$\mathbf{0.95}$ & $0.97$ & $\mathbf{0.96}$&$\mathbf{0.85}$ & $\mathbf{0.83}$ & $\mathbf{0.84}$&$\mathbf{0.85}$ & $0.87$ & $\mathbf{0.86}$&$\mathbf{0.87}$&$0.06$\\
        \hline
    \end{tabular}
}
\end{table*}

By generality, we mean the ability to maintain a consistent performance across different types of data. For this purpose, we use the \textit{coefficient of variation} ({CV}) as the metric, which is defined as the standard deviation divided by the mean of a principal metric, chosen as the  $F_{1}$-score in our case.\par
The latency refers to the time taken from the model under consideration receives an input sample till the anomaly score is generated.\par 
The device resource utilization is measured using an open-source tool {\it s-tui} \cite{manuskintui}, where the CPU and memory utilization for the processes executing the code for the anomaly detection model(s), are measured separately.\par 
The power consumption is measured using a wall ammeter, in Watt (W). The power plug of the device board is plugged into the ammeter, which in turn is plugged into the wall power socket. By power consumption, we mean the increase in power consumption (in \%) when the code for a model is run against an established baseline (the idle state running power), with an LED monitor, keyboard, mouse, and a USB ethernet, connected to the device board, as seen in \fref{fig:odroid}. 
The idle state running power of the experimental setup  was found to be $29.8$ Watt.

{\bf New metric.}
We also propose a new, comprehensive metric called \textit{ADCompScore}, which summarizes all the above metrics into a single quantity, in order to provide a convenient measure of the overall performance of an anomaly detection model, for quick decision-making regarding the algorithm's deployability on an edge device. It is defined as
\begin{multline}\label{eq:ps}
\hspace{-4mm}ADCS = \frac{1}{\sum w_*} \Big[ w_{f}\cdot f + w_{g}\cdot (1-g) + w_{l}\cdot (1-l) + \\ w_{c}\cdot (1-c) + w_{r}\cdot (1-r) + w_{p}\cdot (1-p) \Big]
\end{multline}
where $f$ is the $F_{1}$-score, $g$ is the model's coefficient of variation (CV), $l$ is the \textit{min-max} normalized latency, $c$ is the CPU utilization (\%), $r$ is the RAM utilization (\%), $p$ is the increase in power consumption (\%), and $w_* = \{w_{f}, w_{l}, w_{c}, w_{r}, w_{p}, w_{g}\}$ are the weights associated with the corresponding performance measures. The {\em ADCompScore}, or {\em ADCS} in short, is in the range of $[0, 1]$.

Except for the $F_{1}$-score, all the above performance measures are desired to be as low as possible in magnitude. Hence, we use the complement of those performance measures in our \textit{ADCompScore} metric definition. The weights can be specified to prioritize different performance measures that tailor to different applications or requirements.
The \textit{ADCompScore} metric is defined in this general manner such that it can be easily applied to a wide range of scenarios. In our experiment, we have chosen to simulate the scenario where the different performance metrics are weighted equally, as this allows us to better understand the contribution of each individual performance metric towards the overall score. However, the design of this new metric incorporates flexibility by allowing different weights to be assigned depending on the actual situation at hand.

\subsection{Results}\label{subsec:result}
\subsubsection{Quantitative Performance}

The comparison of the proposed approach against the baseline models, on precision and recall, which are subsumed by $F_{1}$-score, is provided in \tref{tab:perf}. As we can observe from \tref{tab:perf}, LightESD-$1$ and LightESD-$2$ outperform all the other baseline approaches in terms of anomaly detection performance. The ability of LightESD to remove multiple seasonalities (if present) and the trend, results in a much better extraction of the residual component which in turn allows for superior anomaly detection. In a nutshell, LightESD implements a simple yet effective approach to detect anomalies. Note that in the ``Generality" column, a larger mean and a smaller CV are desired. The BRVFL-AE\cite{odiathevar2022bayesian} approach comes closest to our proposed model, in terms of the detection performance, followed by ONLAD\cite{tsukada2020neural}. One of the reasons for such performance of these neural network-based models is that, adding more number of hidden layers may have a positive impact on their detection performances, but increasing the number of hidden layers will add to the training complexity, along with a significant increase in resource utilization and power consumption, which are not desirable for edge applications.  

\begin{table*}[ht]
  \centering
  \caption{Performance Comparison on Additional Factors on SBC (single-board computer).}
  \label{tab:perf_power}
\resizebox{1.0\linewidth}{!}{
      \begin{tabular}{|c|c|c|c|c|}
        \hline
        \textbf{Models}&\textbf{Latency (s)}&\textbf{CPU Utilization (\%)}&\textbf{RAM Utilization (\%)}&\textbf{Increase in Power Consumption (\%)}\\
        \hline
        OC-SVM & $1.34$ & $8.35$ &$3.87$&$23.8$\\
        \hline
        LOF & $0.4$ & $3.6$ & $0.01$&$14.7$\\
        \hline
        Iso. For. & $0.25$ & $3.3$ & $0.01$&$14.5$ \\
        \hline
        ONLAD & $0.19$ & $8.87$ & $3.11$&$16.3$\\
        \hline
        BRVFL-AE & $0.21$ & $10.96$ & $5.33$ & $17.7$\\
        \hline
        
        \rowcolor{yellow!20}
        LightESD-1 & $0.24$ & $5.47$ & $3.29$&$\mathbf{14.3}$\\
        \hline
        \rowcolor{green!20}
        LightESD-2 & $0.24$ & $5.47$ & $3.29$&$\mathbf{14.3}$ \\
        \hline
    \end{tabular}
}
\end{table*}
\begin{figure}[ht]
    \centering
  {\includegraphics[width=0.9\linewidth]{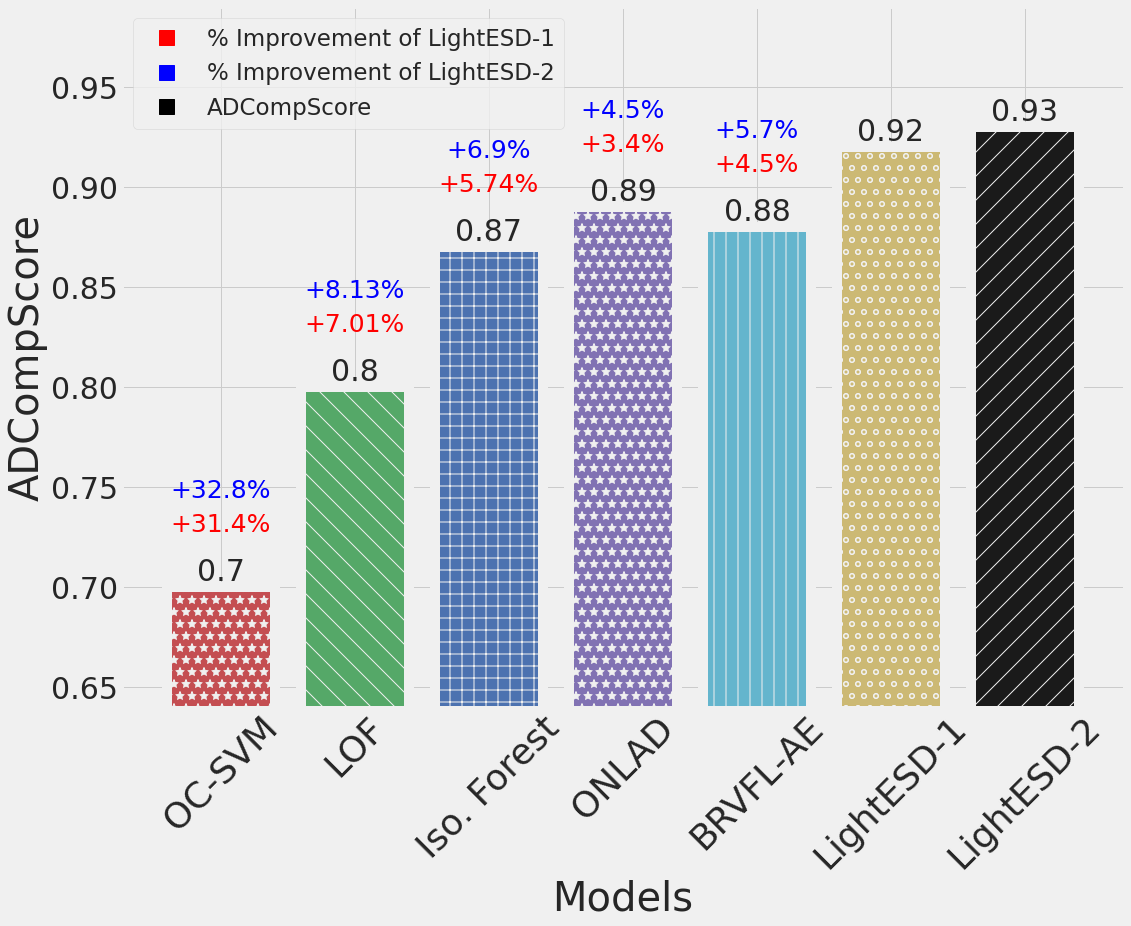}}
  \caption{\textit{ADCompScore} of the different models.}
  \label{fig:ps}
\end{figure}

\tref{tab:perf_power} gives the performance on other metrics which are particularly relevant to edge computing. We see that the proposed approach, LightESD (both LightESD-$1$, and LightESD-$2$), performs well when compared with the other baseline methods. Note that LightESD does not need to store any weights/coefficients for the inference stage which is unlike the other methods, and this results in a low RAM utilization especially when compared to SOTA methods, as well as the lowest power consumption. SOTA approaches like ONLAD and BRVFL-AE, need to store their \textit{weight matrices} and \textit{bias vectors}, thereby adding extra overhead to memory utilization. However, observing \tref{tab:perf_power}, we also see that both ONLAD and BRVFL-AE have better performance in terms of latency, and our proposed approach falls short against the SOTA methods possibly due to the iterative nature of the proposed LightESD algorithm. While ML-methods like the Local Outlier Factor and Isolation Forest have very competitive performances in terms of CPU and RAM utilization, they fall short in terms of anomaly detection performance.

As we have observed so far, a model cannot perform in a superior manner in {\em all} aspects. In order to have a holistic view of the overall performance of all the models, we utilize the \textit{ADCompScore} evaluation metric as defined in Eq. \eqref{eq:ps}.  \fref{fig:ps} provides the overall performance of the proposed LightESD approach for both LightESD-$1$ and LightESD-$2$, as well as the baseline methods, using our proposed {\em ADCompScore} evaluation metric, where all the weights $w_*$ are set to 1. We can see that, although BRVFL-AE \cite{odiathevar2022bayesian} has a lower latency (as seen in \tref{tab:perf_power}) as well as a good detection performance, it falls short in terms of the \textit{ADCompScore} due to its much higher device resource utilization. Along similar lines, machine learning baseline methods like Isolation Forest and the Local Outlier Factor have very competitive device resource utilization and low power consumption, but fall short in their detection performance. \fref{fig:ps} also quantifies the percentage improvement of the proposed, LightESD, anomaly detection model with respect to all the baseline methods, in terms of the overall {\em ADCompScore}. The proposed LightESD approach is able to beat the SOTA models (BRVFL-AE and ONLAD) by a margin of $3 - 6\%$, and the other ML-based models by more significant margins (up to $33\%$). Taking into account all these factors besides the anomaly detection performance, which are of great importance to rapid, on-device edge learning and deployment, LightESD is able to exceed the overall performance of the SOTA edge anomaly detectors (like BRVFL-AE and ONLAD), as well as popularly used ML-based detectors, thereby making it a much desirable anomaly detection method for Edge AI applications.

\subsubsection{Effect of Significance Level}\label{sec:sig}
As we observe in Table \ref{tab:perf}, decreasing the significance level, $\alpha$, from $0.05$ to $0.001$, significantly increases the \textbf{Precision} of LightESD, without notably impacting the \textbf{Recall}. In other words, with this change, the proposed model is able to reduce the number of \textit{false positives} (a.k.a. \textit{false alarms}) without negatively impacting the number of \textit{false negatives} (i.e., \textit{missed detection} of true anomalies), which is desirable for different kinds of problems.

\section{Conclusion}
This paper proposes an anomaly detection framework that is able to detect anomalies directly at the edge site, without the need for any training at a central server. Unlike many statistical and machine learning methods, it is non-parametric and does not require any assumption of the underlying distribution of input data. It is also weight-free and does not require separate training and validation phases, as the model auto-fits/adapts to the underlying data directly on the fly to identify outlier data points. 
With a focus on time series applications, the proposed approach, LightESD, can tackle different types of data including seasonal, non-seasonal, and random walks. Our comprehensive evaluation demonstrates that LightESD outperforms both SOTA and other popular anomaly detectors by clear margins, as well as mitigates false alarms. It also generalizes much better across different datasets both on synthetic and real-world data. Moreover, LightESD consumes very low power and CPU/memory resources, when compared to SOTA anomaly detection schemes for the edge. These attributes make LightESD a desirable choice for rapid deployment directly at the edge, and for producing near real-time detection performance due to the low latency.

One limitation of LightESD is that it requires a batch of data for learning the underlying patterns, which would face challenge when there is only a single training instance in some circumstances. Thus in future work, we plan to enhance LightESD from a {\em batch learning} setting to a pure {\em online learning} setting.

\bibliographystyle{IEEEtran}
\bibliography{IEEEabrv,refs}

\end{document}